# Tracking Particles Ejected From Active Asteroid Bennu With Event-Based Vision

Loïc James Azzalini[1,a] *, Dario Izzo[1,b]

[1] Advanced Concepts Team, European Space Agency, European Space Research and Technology Centre (ESTEC), Keplerlaan 1, 2201 AZ Noordwijk, The Netherlands

[a]jazzalin@outlook.com, [b]dario.izzo@esa.int

**Abstract.** Early detection and tracking of ejecta in the vicinity of small solar system bodies is crucial to guarantee spacecraft safety and support scientific observation. During the visit of active asteroid Bennu, the OSIRIS-REx spacecraft relied on the analysis of images captured by onboard navigation cameras to detect particle ejection events, which ultimately became one of the mission's scientific highlights. To increase the scientific return of similar time-constrained missions, this work proposes an event-based solution that is dedicated to the detection and tracking of centimetre-sized particles. Unlike a standard frame-based camera, the pixels of an event-based camera independently trigger events indicating whether the scene brightness has increased or decreased at that time and location in the sensor plane. As a result of the sparse and asynchronous spatiotemporal output, event cameras combine very high dynamic range and temporal resolution with low-power consumption, which could complement existing onboard imaging techniques. This paper motivates the use of a scientific event camera by reconstructing the particle ejection episodes reported by the OSIRIS-REx mission in a photorealistic scene generator and in turn, simulating event-based observations. The resulting streams of spatiotemporal data support future work on event-based multi-object tracking.

**Introduction**
An active asteroid shows evidence of mass loss caused by natural processes or as a result of human-planned activities, sharing characteristics more often associated with comets. Detecting and interpreting the dynamic properties of these small solar system bodies (SSSB) constitute important scientific objectives of sample-return and flyby missions as they may hold the key to understanding their past and future [1]. In the case of asteroid Bennu, such activity was only detected in situ as the navigation cameras of visiting spacecraft OSIRIS-REx captured centimetre-size rocks being ejected from the surface. Whether for situational awareness to guarantee spacecraft safety or scientific observation, missions to active asteroids or comets benefit from early detection and tracking of such events [2]. Currently, extensive offline image analysis is needed to balance the brightness of the body and the dimness of the particles (stray light reduction), to negate the static objects in the scene (image differencing) and to detect the relative motion of the particles (blinking) [1]. Given the time-constrained nature of these missions, techniques that automate the detection and tracking of particles, such as the frame-based multi-object tracking algorithm proposed in [3], could significantly contribute to the mission's scientific returns.

This work proposes a solution based on the principles of dynamic vision sensing and the event camera, a novel imaging sensor inspired by the neural pathways of the retina [4, 5]. Where a standard frame-based camera would capture redundant static information, an event camera would only report pixel-level brightness changes induced by the dynamics present in the scene. The sparse and asynchronous output of the sensor and its large dynamic range provide an effective low-power solution to detection and tracking problems in challenging lighting conditions [6]. Given particularly dynamic environments around SSSBs as a result of plumes, dust and/or particle ejecta, we theorise that event-based technology could augment onboard navigation and scientific cameras, streamlining current approaches based on offline image analysis.

The apparent advantages of these dynamic vision sensors for space applications have recently been reviewed in [7]. While the principles of dynamic vision have yet to be applied to the detection and tracking of particles around an active asteroid, similar work on event-based star tracking [8] and (event-based) space situational awareness [9, 10] provide valuable insights into the challenges posed by the problem at hand. Unlike the streaks produced by resident space objects (RSOs) crossing the field of view of a ground-based or in-orbit telescope, the particles of interest in this study undergo more complex dynamics locally, given the proximity of the SSSB, limiting the utility of classical approaches such as the Hough transforms. Moreover, the problem of associating observations to tracks is exacerbated by significant measurement noise that is inherent to the event camera. Popular multi-object tracking solutions such as the (probabilistic) multiple hypothesis tracking filters have been adapted to event-based representations of RSOs [9, 11], suggesting that these algorithms also lend themselves to the analysis of particle dynamics in the vicinity of an SSSB.

To support the evaluation of event-based multi-particle tracking, we use the navigation and ancillary information of the OSIRIS-REx mission in simulation as well as reports of notable particle ejection episodes [12] to highlight how an event-based sensor can augment visual data capture and contribute to the mission's scientific objectives. The dynamic scene, composed of asteroid Bennu and several centimetre-size particles ejected from its surface, is first reconstructed with photorealistic computer graphics tools from the point of view of the visiting spacecraft [13]. Dynamic vision sensing is subsequently simulated by emulating the sensitivity and noise characteristics of a real event camera [14, 15]. In turn, we demonstrate that, in the absence of frames, asynchronous streams of events capture sufficient information about the scene to enable fast and continuous tracking of multiple particles. This constitutes an essential step to support future research on event-based determination of the size and orbit of RSOs.

The following section introduces the principle of operation underlying the high temporal resolution and high dynamic range of the event camera. Section Simulation and Results describes the simulation environment in which the dynamic scene is reproduced and the preliminary results on event-based tracking of particle ejecta. Given the sparse and asynchronous output of the event camera, future work will focus on the formulation of a multi-object tracking problem that is compatible with the event-based representation of the particle tracks.

**Theoretical Background**

Event-Based Vision. The pixels of a dynamic vision sensor independently report changes in scene brightness, which is generally taken to be the log intensity (i.e., $log\ I$), resulting in a sparse and asynchronous stream of *events*. The pixel at location $(x, y)$ in the sensor plane outputs a 1 (alternatively, 0) if the sensed brightness at time $t$ increased (decreased) by the predefined magnitude $\Theta_{ON}$ ($\Theta_{OFF}$) known as the contrast thresholds. Thus, an event can be represented by the 4-tuple $(t, x, y, p)$, where $p$ denotes the polarity of the brightness change. After triggering an event, the pixel resets its baseline from which to monitor subsequent changes in brightness to the current log intensity, as depicted in Figure 1 (a). Figure 1 (b) shows a synthetic example of a stream of events output by an event camera capturing a black dot spinning on a white disk. As only the motion of the black dot contributes changes in brightness in the scene, the spatiotemporal output of the sensor takes the form of a spiral.

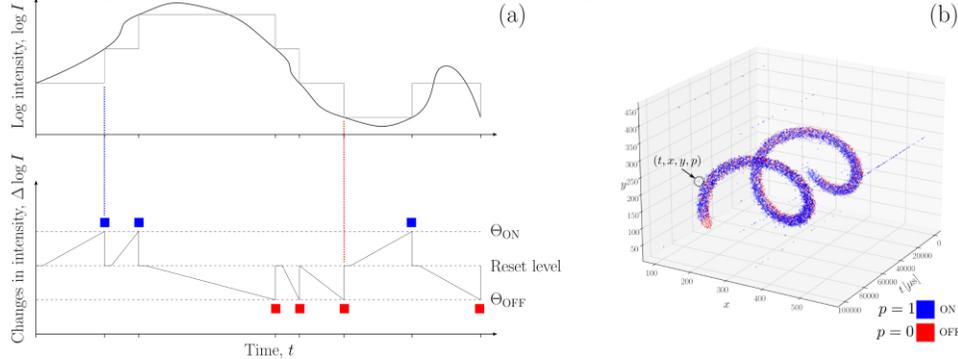

**Figure 1** Event-based vision sensing overview: (a) principle of operation of a dynamic vision pixel (adapted from [4]) (b) example output from an event camera capturing a black dot spinning on a white disk (based on synthetic data from the Metavision toolkit [16])

The sparsity of the output allows for high readout rates in the range of $2 - 1200\ MHz$, high temporal resolution (in the order of µs), low-power consumption and, owing to the logarithmic scale used by each pixel, high dynamic range [5]. While these are appealing characteristics for an onboard sensor, the event camera is inherently noisy due to the complex transistor circuits underlying its differential mode of operation. Figure 1 (b) captures some of these noisy contributions in the form of sporadic streaks of ON events at several pixel locations in the sensor plane. These artefacts are the result of nonidealities such as hot pixels and leak currents which produce events irrespective of the scene dynamics. In order to faithfully simulate the output of these sensors, event camera emulators [14, 15] model these sources of noise and other hardware-related nonidealities (e.g., shot noise).

**Simulation and Results**

The first step of the simulation of dynamic vision sensing around active asteroid Bennu consists in extracting the relative position of the OSIRIS-REx spacecraft, the Sun and the ejected particles. The Orbital C mission phase is of particular interest given that it was dedicated to particle monitoring, resulting in a higher cadence of imaging. Specifically, a time window centred on the particle ejection episode of September 13[th], 2019, is considered in this work given that out of the 30 particles detected, 22 tracks led to successful orbit determination [12]. Figure 2 depicts the reconstructed particle ejection episode based on the interpolation of the relevant SPICE kernels.

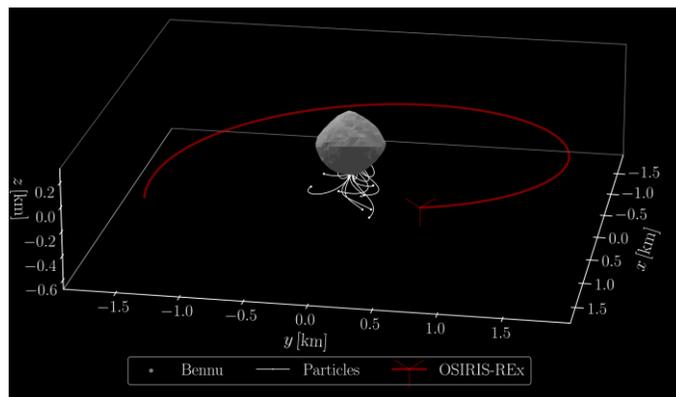

**Figure 2** Bennu-fixed particle ejection visualization based on the interpolation of SPICE kernels from 2019-09-13T21:00:00 to 2019-09-14T00:00:00 [12]

Given the dynamic nature of the event camera, the ejection scene depicted in Figure 2 is then animated in Blender, using photorealistic models of the asteroid and the particles. Different textures and reflectance properties are used to reproduce the rubble pile appearance of Bennu [13] based on the position of the Sun during the Orbital C phase. The simulated ejecta consists of 14 particles with sizes ranging from $1 - 11\ cm$ according to the average diameters reported in [1]. We emulate the large field of view and image sensor size of the onboard navigation camera (NavCam 1) by configuring the Blender perspective camera with $W = 2592$ px, $H = 1944$ px and a horizontal field of view of $\vartheta = 44°$. Then, the positions of the particles and the emulated camera along the Orbital C trajectory are updated to capture frames of the ejection episode, as shown in Figure 3 (a) (cropped near the limb where the ejecta emanate).

Overall, 40 additional renders are generated and combined into a video at 30 $FPS$ tracing the path of the particles from the surface of Bennu to the lower-left borders of the field of view. To study an event-based representation of the scene dynamics, the video is then passed to an event camera emulator [14, 15], which converts the video to spatiotemporal streams of events based on the operation principle described in Section Event-Based Vision. In the interest of qualitative comparison with the original renders, synthetic events are binned into event-frames over short time windows such that the original frame rate is recovered.

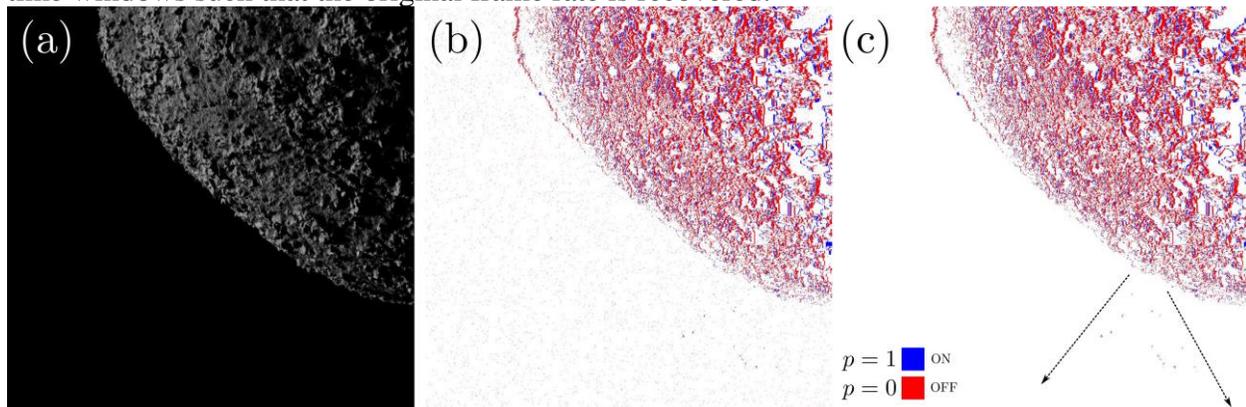

**Figure 3** Reconstruction of a particle ejection episode: (a) the particles are difficult to detect in the photorealistic render, (b) yet accumulation of synthetic events in a single frame make the particles more easily identifiable against the dynamic background noise (b) and clearly visible in the absence of noise (c) (the arrows indicate the ejecta direction)

While the low-light conditions depicted in Figure 3 (a) significantly challenge the detection of any particles, the accumulation of synthetic events depicted in (b) allows the particles to be visually distinguished from the asteroid, despite significant noise contributions from the sensor emulation (to aid in locating the particles, frame (c) depicts a noiseless emulation of the same event camera).

**Conclusion**

This work showcases preliminary results on the scientific use of event cameras onboard a spacecraft in dynamic environments around small solar system bodies. After reconstructing the scene of a particle ejection episode from the SPICE kernels of the OSIRIS-REx mission, we emulate the viewpoint of an event camera to gain familiarity with event-based representations of particle tracks. In follow-up work, we will evaluate the formulation of a multi-object tracking problem, given event-based observations of an unknown (and potentially time-varying) number of particles. Multi-object tracking algorithms may be applied directly to the event-frames shown in Figure 3. However, this defeats the purpose of emulating an event camera in the first place, as by so doing, solutions do not take full advantage of the sparsity and temporal resolution of the data. Instead, we will consider approaches that process events in an online manner.

**Acknowledgements**

We would like to thank our colleagues at ESA's Advanced Concepts Team for their ongoing support. LJA acknowledges support through ESA's Young Graduate Trainee (YGT) program.

**References**

[1] Lauretta, D. S. et al. (2019). Episodes of particle ejection from the surface of the active asteroid (101955) Bennu. Science, 366 (6470), https://doi.org/10.1126/science.aay3544

[2] Chesley, S. R. et al. (2020). Trajectory estimation for particles observed in the vicinity of (101955) Bennu. Journal of Geophysical Research: Planets, 125, e2019JE006363. https://doi.org/10.1029/2019JE006363

[3] Liounis, A. J. et al. (2020). Autonomous detection of particles and tracks in optical images. Earth and Space Science, 7, e2019EA000843. https://doi.org/10.1029/2019EA000843

[4] Liu, S.-C., & Delbruck, T. (2010). Neuromorphic sensory systems. Current Opinion in Neuro- biology, 20 (3), 288-295. https://doi.org/10.1016/j.conb.2010.03.007

[5] Gallego, G. et al. (2022). Event-based vision: A survey. IEEE Transactions on Pattern Analysis and Machine Intelligence, 44 (1), 154-180. https://doi.org/10.1109/TPAMI.2020.3008413

[6] Lagorce, X. et al. (2015). Asynchronous Event-Based Multikernel Algorithm for High-Speed Visual Features Tracking. IEEE Transactions on Neural Networks and Learning Systems, vol. 26, no. 8, pp. 1710-1720. https://doi.org/10.1109/TNNLS.2014.2352401

[7] Izzo, D. et al. (2022). Neuromorphic computing and sensing in space. arXiv preprint arXiv:2212.05236. https://doi.org/10.48550/arXiv.2212.05236

[8] Chin, T.-J., et al. (2019). Star tracking using an event camera. In 2019 IEEE/CVF Conference on Computer Vision and Pattern Recognition Workshops (CVPRW) (p. 1646-1655). https://doi.org/10.1109/CVPRW.2019.00208

[9] Cheung, B. et al. (2018). Probabilistic multi hypothesis tracker for an event based sensor. In 2018 21st International Conference on Information Fusion (fusion) (p. 1-8). https://doi.org/10.23919/ICIF.2018.8455718

[10] Afshar, S. et al. (2020). Event-Based Object Detection and Tracking for Space Situational Awareness. IEEE Sensors Journal, vol. 20, no. 24, pp. 15117-15132. https://doi.org/10.1109/JSEN.2020.3009687

[11] Oliver, R. et al. (2022). Event-based sensor multiple hypothesis tracker for space domain awareness. In AMOS Conference 2022. https://doi.org/10.5167/uzh-231276

[12] Hergenrother, C. W. et al. (2020). Photometry of particles ejected from active asteroid (101955) Bennu. Journal of Geophysical Research: Planets, 125, e2020JE006381. https://doi.org/10.1029/2020JE006381

[13] Pajusalu M. et al. (2022) SISPO: Space Imaging Simulator for Proximity Operations. PLOS ONE 17(3): e0263882. https://doi.org/10.1371/journal.pone.0263882

[14] Gehrig, D. et al. (2020). Video to events: Recycling video datasets for event cameras. In 2020 IEEE/CVF Conference on Computer Vision and Pattern Recognition (CVPR) (p. 3583-3592). https://doi.org/10.1109/CVPR42600.2020 .00364

[15] Hu, Y. et al. (2021). v2e: From Video Frames to Realistic DVS Events. 2021 IEEE/CVF Conference on Computer Vision and Pattern Recognition Workshops (CVPRW), Nashville, TN, USA, 2021, pp. 1312-1321, https://doi.org/10.1109/CVPRW53098.2021.00144

[16] Prophesee.ai. Metavision SDK Docs – Recordings and Datasets, https://docs.prophesee.ai/